\documentclass[10pt, a4paper]{article}

\usepackage[final]{lrec2026} 
\usepackage{graphicx}
\usepackage{amsmath,amssymb} 
\usepackage{color}
\usepackage{kotex}
\usepackage{caption}

\usepackage{booktabs}
\usepackage{multirow}
\usepackage{wrapfig}
\usepackage{tabularx}
\usepackage{booktabs}
\usepackage{array}    

\title{Team MKC at MEDIQA-SYNUR 2026: Retrieval-Augmented Generation Based Nurse Observation Extraction}

\name{Kyomin Hwang\textsuperscript{1}, Nojun Kwak\textsuperscript{1,2\dag}\thanks{\dag\ \ Corresponding author.}}

\address{\textsuperscript{1}GSCST, Seoul National University, \textsuperscript{2}AIIS, Seoul National University \\
         \{kyomin98, nojunk\}@snu.ac.kr\\}

\abstract{
Recent advancements in Large Language Models (LLMs) have played a significant role in reducing human workload across various domains, a trend that is increasingly extending into the medical field. In this paper, we propose an automated pipeline designed to alleviate the burden on nurses by automatically extracting clinical observations from nurse dictations. To ensure accurate extraction, we introduce a method based on Retrieval-Augmented Generation (RAG). Our approach demonstrates effective performance, achieving an F1-score of 0.796 on the MEDIQA-SYNUR test dataset. \\ \newline  }

\begin{document}

\maketitleabstract

\section{Introduction}

Recent advancements in Large Language Models (LLMs), exemplified by the GPT series~\cite{gpt}, have accelerated a paradigm shift across a multitude of domains, demonstrating remarkable capabilities in natural language understanding and generation. This momentum has profoundly impacted the medical field~\cite{kim-etal-2024-team}, leading to the development of specialized models such as MedGemma~\cite{sellergren2025medgemma} and LLaVA-Med~\cite{li2023llava}. These models have achieved state-of-the-art performance on various biomedical tasks by leveraging large-scale, domain-specific datasets. However, obtaining such high-quality annotated data remains a significant bottleneck due to privacy concerns and the high cost of expert annotation. In this context, the SYNUR~\cite{synur} dataset has established a crucial benchmark, facilitating the automatic extraction of clinical observations from nursing dictations—a task essential for streamlining clinical documentation.

Building upon these developments, we propose a novel Retrieval-Augmented Generation (RAG) pipeline designed to accurately extract clinical observations from complex nurse dictations. Unlike conventional approaches that rely on resource-intensive fine-tuning, our automated pipeline integrates a synergistic dual-retrieval mechanism. First, we employ an ontology-based retrieval system to fetch the most relevant medical concepts based on the current utterance, ensuring terminological precision. Second, we utilize a memory bank to retrieve semantically similar dialogue segments and their corresponding gold-standard observations, enabling the model to learn extraction patterns via in-context learning. This dual approach equips the foundation LLM with both explicit medical context and structural guidance. Utilizing this training-free framework, we achieved an F1-score of $0.796$ on the MEDIQA-SYNUR test set. These results demonstrate that our approach can effectively alleviate the nursing documentation burden by automating the interpretation of patient interactions with high reliability.
\section{Related Works}

\subsection{Foundation model in Medical Domain}

Recent advancements in large-scale data acquisition and the exponential growth of computing power have led to the emergence of foundation models capable of performing diverse tasks within a unified framework~\cite{Jang_Kong_Jeon_Kim_Kwak_2023, gpt}. This paradigm has significantly influenced the medical domain, triggering the development of specialized foundation models. Early efforts focused on encoder-based models like ClinicalBERT~\cite{huang2019clinicalbert} and BlueBERT~\cite{peng2019transfer}, which are optimized for medical knowledge representation and retrieval. More recently, generative and multi-modal models such as LLaVA-Med~\cite{li2023llava} and MedGemma~\cite{sellergren2025medgemma} have been introduced to handle complex biomedical reasoning. Currently, active research is dedicated to adapting these generalist foundation models to enhance their performance on specific clinical tasks~\cite{li2023chatdoctor, toma2023clinical}. Aligning with this research process, this paper proposes a method for developing a Large Language Model (LLM) specifically optimized for analyzing medical observations within nurse dictations.

\subsection{Retrieval-Augmented Generation}

\begin{figure*}[ht]
    \centering
\includegraphics[width=1.0\textwidth]{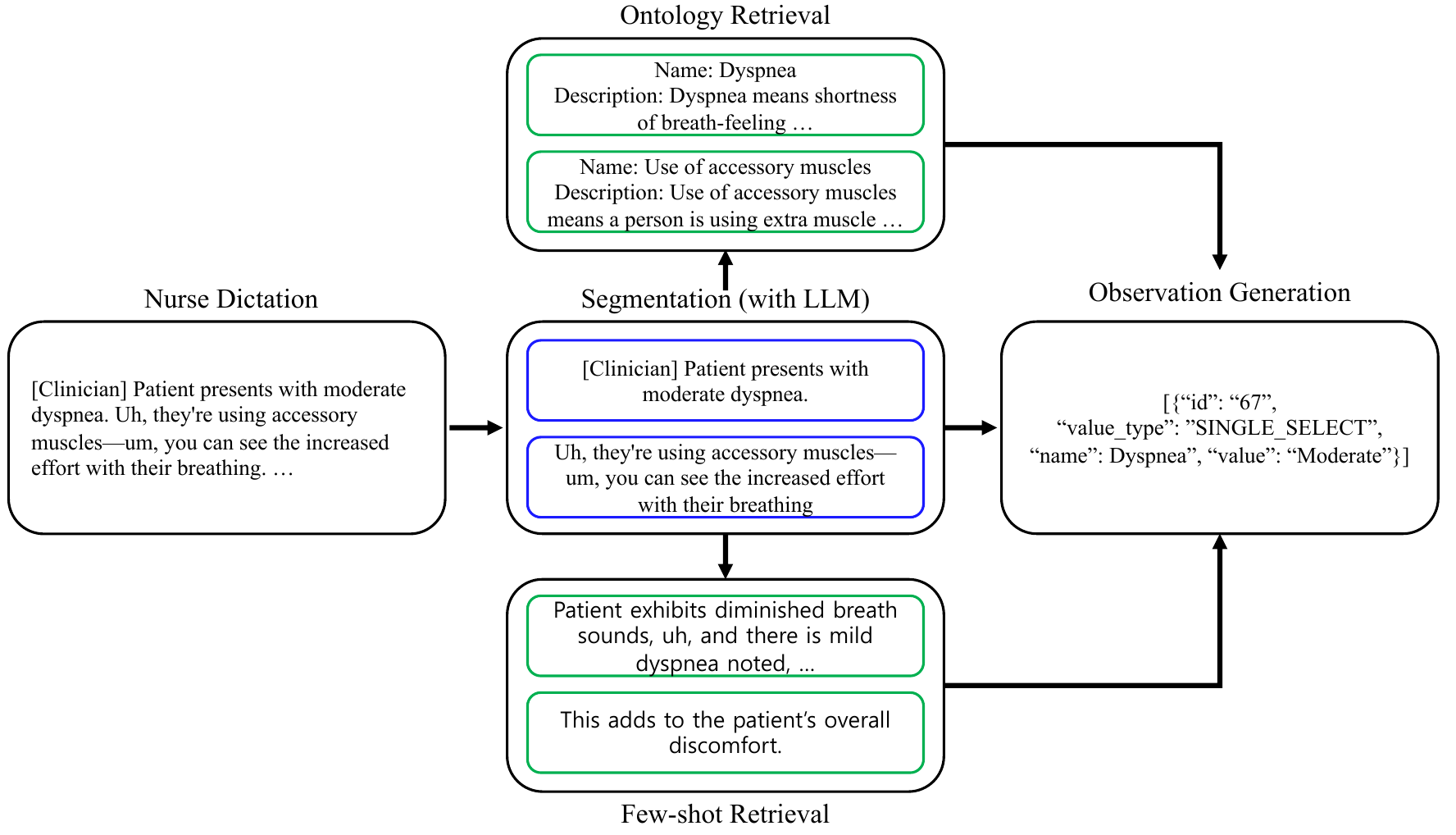}
\caption{Full illustration of Retrieval-Augmented Generation (RAG) Based Nurse Observation Extraction
Pipeline}
     \label{fig:full_fig}
\end{figure*}

Retrieval-Augmented Generation (RAG)~\cite{lewis2020retrieval, kim2024nice} has emerged as a prominent approach for tailoring foundation models to specific tasks. While foundation models possess vast general knowledge, they often struggle with domain-specific nuances and factual accuracy in critical fields like healthcare. RAG addresses this limitation by grounding the model's responses in retrieved evidence. Previous studies have demonstrated that utilizing RAG techniques can significantly enhance task-specific performance across various domains without the need for additional training of the foundation models~\cite{chen2022murag, kim2024nice}. Leveraging this capability to handle complex clinical terminologies and contexts, we propose a RAG-based automatic generation pipeline designed to specialize a foundation LLM for extracting medical observations from nurse dictations, all without requiring further model training.
\section{Method}

\begin{figure*}[ht]
    \centering
\includegraphics[width=1.0\textwidth]{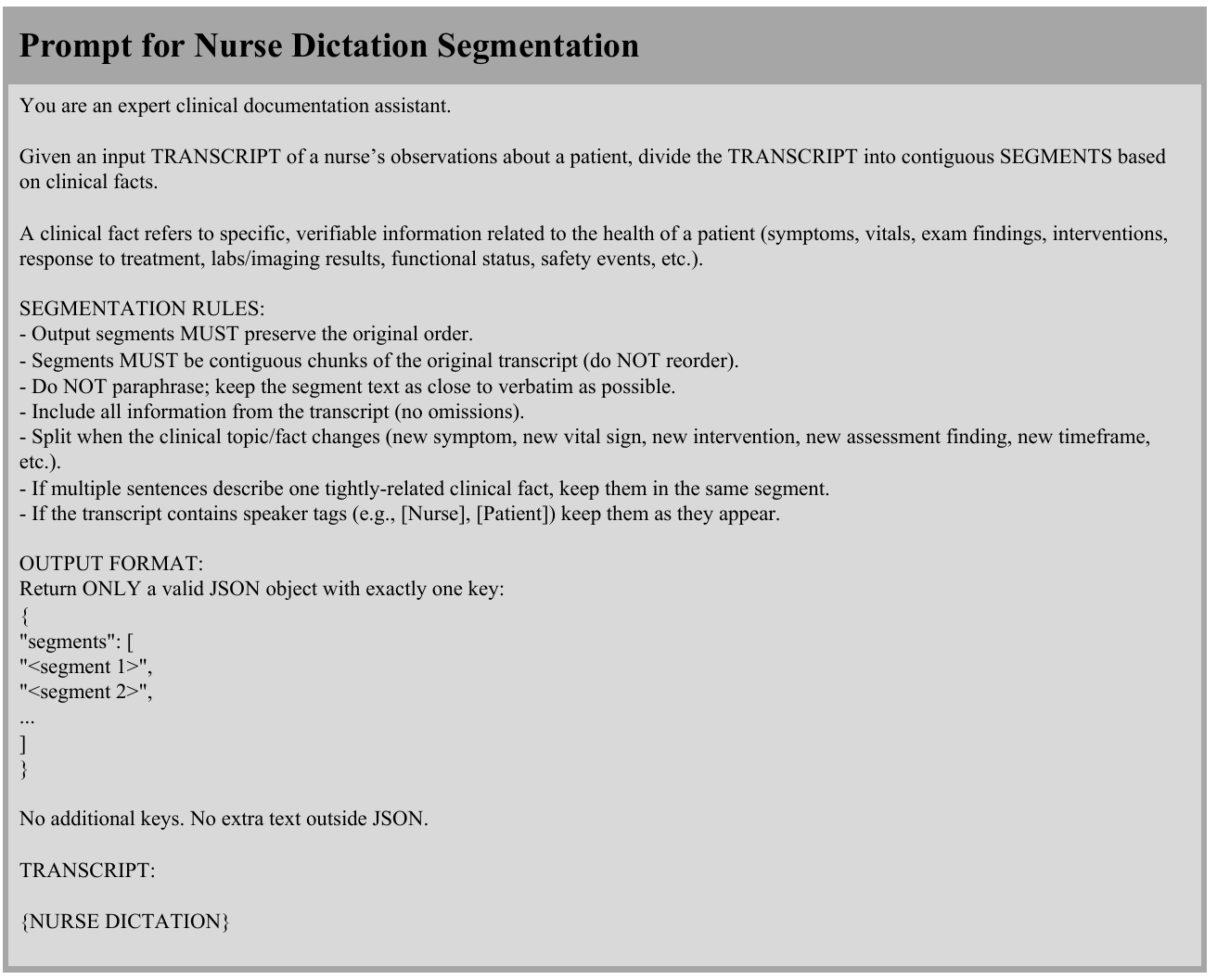}
\caption{Prompt for nurse dictation segmentation}
     \label{fig:segment}
\end{figure*}

In this section, we present a pipeline designed to automatically extract clinical observations from nurse dictations. Figure~\ref{fig:full_fig} illustrates our medical observation pipeline.

\begin{figure*}[ht]
    \centering
\includegraphics[width=1.0\textwidth]{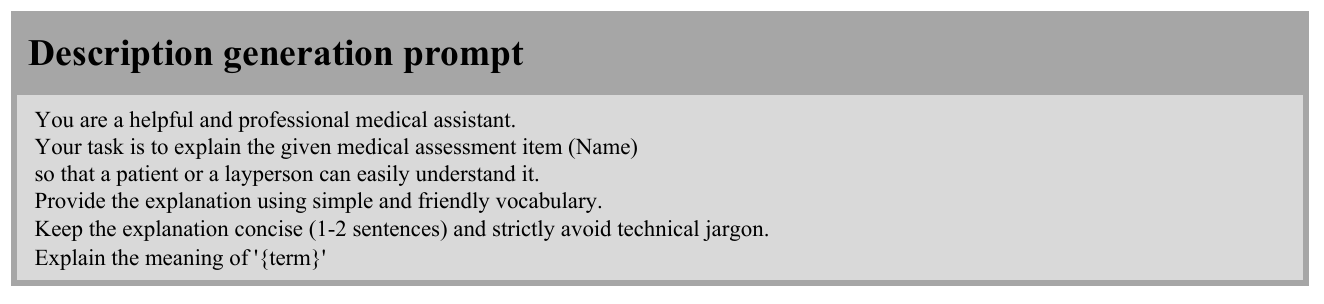}
\caption{Prompt for description generation}
     \label{fig:desc}
\end{figure*}

\subsection{Nurse Dictation Segmentation}

Nurse dictations processed by LLMs are typically lengthy. Therefore, following the approach proposed by \cite{synur}, we first segment the input dictation into distinct units containing clinical facts. The prompt used for this segmentation is shown in Figure~\ref{fig:segment}.

Each nurse dictation is divided into multiple segments. For each segment, we generate observations by providing the LLM with two types of retrieved context: 1) relevant schemas retrieved from the ontology, and 2) semantically similar segments retrieved from the training dataset. The latter serves as few-shot examples, consisting of segment-observation pairs, to guide the LLM in generating the correct observation for the input segment.

\begin{figure*}[ht]
    \centering
\includegraphics[width=1.0\textwidth]{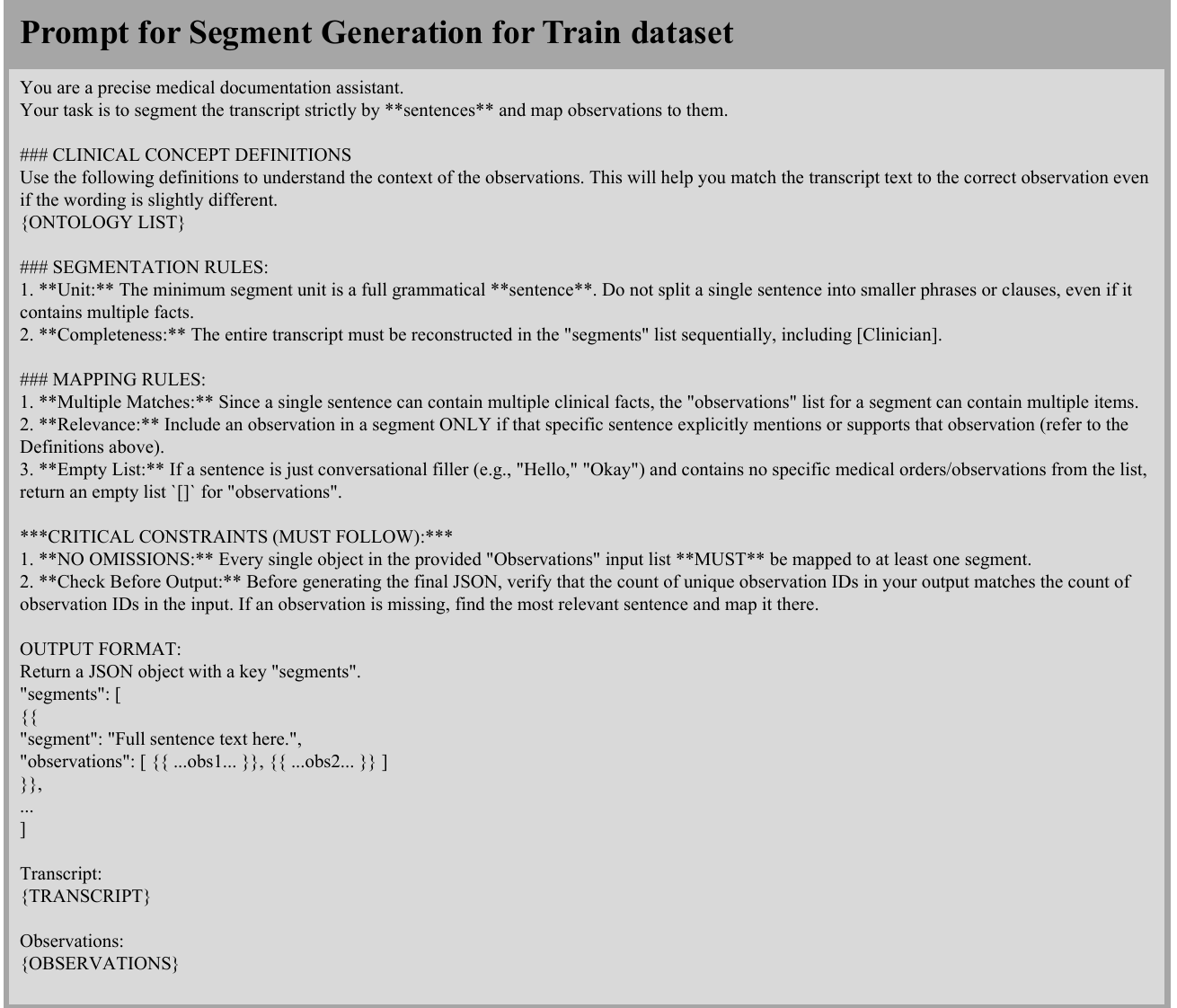}
\caption{Prompt for segment generation for train dataset}
     \label{fig:segment_train}
\end{figure*}

\begin{figure*}[ht]
    \centering
\includegraphics[width=1.0\textwidth]{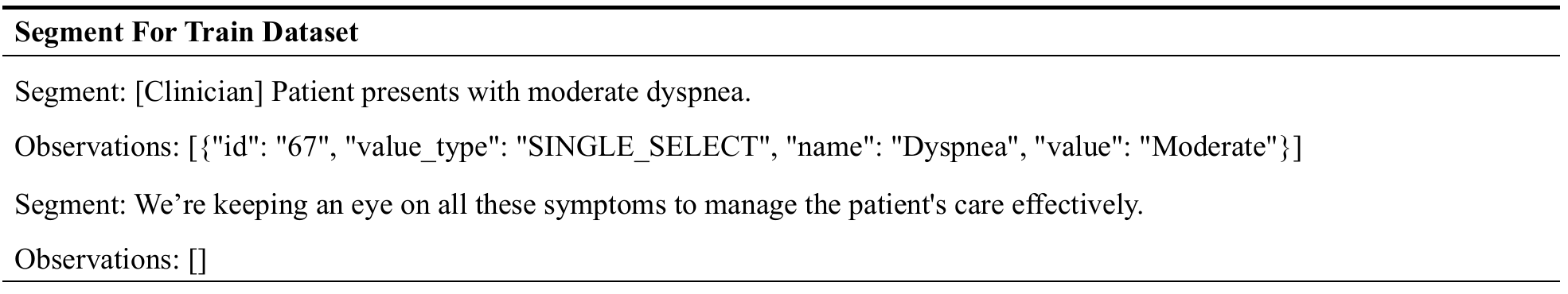}
\caption{Segment Example for Train Dataset.}
     \label{fig:segment_ex}
\end{figure*}

\begin{figure*}[ht]
    \centering
\includegraphics[width=1.0\textwidth]{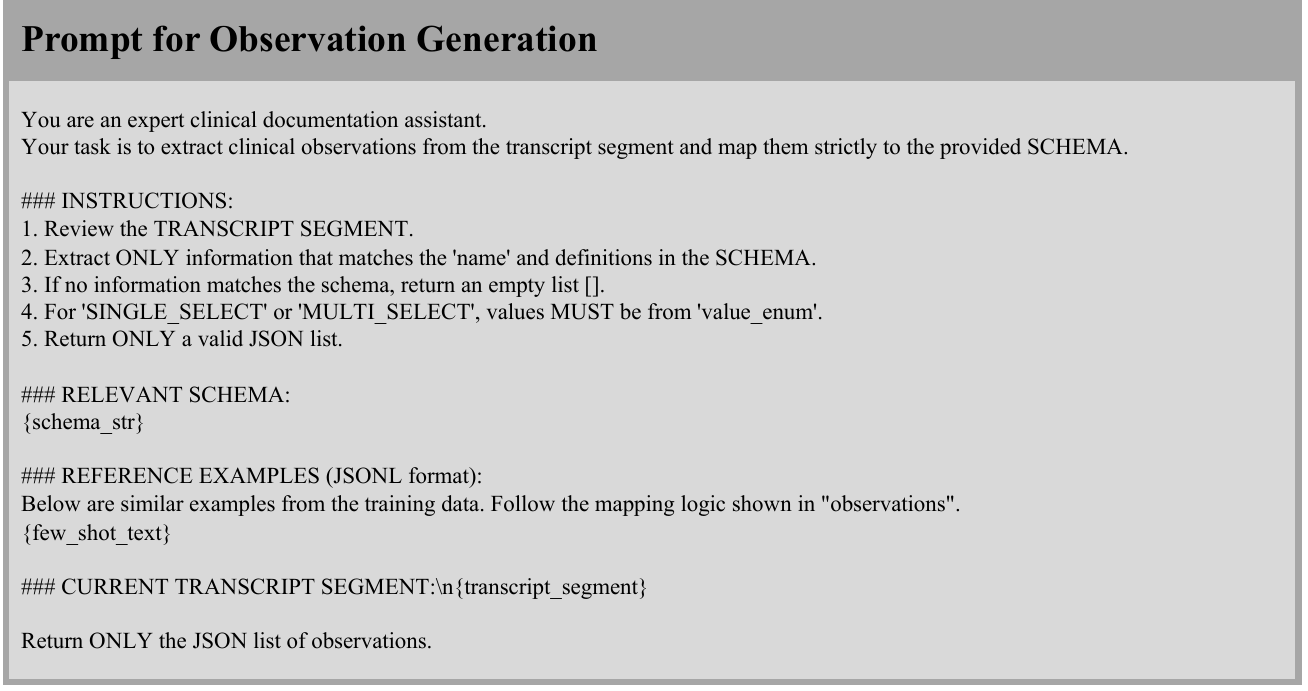}
\caption{Prompt for observation generation}
     \label{fig:obser_gen}
\end{figure*}

\subsection{Description for Medical Observation in Ontology}

First, we recognized that directly inputting medical observations defined within an ontology (e.g., weight-bearing status) into an LLM could lead to performance degradation. Consequently, we implemented a process to append descriptions to each medical observation. Figure~\ref{fig:desc} illustrates the prompt used to generate these descriptions.

\subsection{Retrieving Relevant Schemas from Ontology}

To enhance the accuracy of observation extraction, we identify schemas relevant to the input segment from the ontology and provide them as input to the LLM. For retrieval, we employ a hybrid approach utilizing both BlueBert~\cite{peng2019transfer} and TF-IDF. We format each schema using its name, description and its value enumerations (options). The input format is as follows:

\begin{align} \text{{name}. {Description} Options: {Option1}, {Option2}} \end{align}

We select the top 10 most relevant schemas to include in the LLM input.

\subsection{Retrieving Few-Shot Examples from Training Dataset}

In addition to schemas, we utilize few-shot examples to improve generation performance by retrieving segments from the training dataset that are similar to the input segment, along with their ground-truth observations.

To construct the retrieval pool, we first processed the training dataset using ChatGPT. We extracted individual segments and their corresponding observations to create a database of pairs. Figure~\ref{fig:segment_train} illustrates the prompt used for this extraction process, while Figure~\ref{fig:segment_ex} displays examples of the resulting segment-observation pairs. From this processed dataset, we retrieve the top 15 examples to serve as few-shot context for the LLM. This retrieval step is performed using a hybrid method combining BlueBert and BM25.

\subsection{Observation Generation}

We retrieved $N$ and $K$ entries, respectively, from the ontology pool and the few-shot example pool generated in the previous section, instructing the LLM to extract medical observations relevant to the current utterance. Figure~\ref{fig:obser_gen} presents the prompt used for this task.
\section{Experiments}

\begin{table}[ht]
    \centering
    \caption{Performance comparison of the proposed methods. Best results are marked in \textbf{bold}.}
    \label{tab:main_results}
    \renewcommand{\arraystretch}{1.3}
    \begin{tabularx}{\columnwidth}{X c c c} 
        \toprule
        \textbf{Method} & \textbf{Precision} & \textbf{Recall} & \textbf{F1} \\
        \midrule
        + Few-shot Ex & 0.789 & 0.694 & 0.739 \\
        + Few-shot Ex + Schema & \textbf{0.812} & \textbf{0.847} & \textbf{0.829} \\
        \bottomrule
    \end{tabularx}
\end{table}

\begin{table*}[t] 
    \centering
    \caption{Performance comparison with different configurations on the Test Dataset. \textdagger denotes the usage of refined prompt.}
    \label{tab:test_ex}
    \renewcommand{\arraystretch}{1.2} 
    \setlength{\tabcolsep}{10pt} 
    
    \begin{tabular}{l c c c c c c} 
        \toprule
        \multirow{2}{*}{\textbf{Base Model}} & \multirow{2}{*}{\textbf{Shots}} & \multicolumn{2}{c}{\textbf{LLM Backbone}} & \multicolumn{3}{c}{\textbf{Metrics}} \\
        \cmidrule(lr){3-4} \cmidrule(lr){5-7}
        & & \textbf{Generator} & \textbf{Segmenter} & \textbf{Precision} & \textbf{Recall} & \textbf{F1} \\
        \midrule
                
        \multirow{7}{*}{BlueBert} 
        & 3 & GPT-5-mini & GPT-5-mini & 0.739 & 0.810 & 0.773 \\
        & 5 & GPT-5-mini & GPT-5-mini & 0.744 & 0.817 & 0.779 \\
        & 10 & GPT-5-mini & GPT-5-mini & 0.744 & 0.818 & 0.779 \\
        & 10 & GPT-5.1 & GPT-5-mini & 0.785 & 0.805 & 0.795 \\
        & 10 & GPT-5.2 & GPT-5-mini & 0.766 & 0.825 & 0.794 \\
        & 10 & GPT-5.2 & GPT-5.2 & 0.759 & \textbf{0.827} & 0.792 \\
        & 15 & GPT-5.1 & GPT-5-mini & \textbf{0.786} & 0.807 & \textbf{0.796} \\        
        \bottomrule
    \end{tabular}
\end{table*}

\begin{table}[ht]
    \centering
    \caption{Ablation on retrieval model.}
    \label{tab:abl}
    \renewcommand{\arraystretch}{1.2}
    \begin{tabularx}{\columnwidth}{X c c c} 
        \toprule
        \textbf{Method} & \textbf{Precision} & \textbf{Recall} & \textbf{F1} \\
        \midrule
        ClinicalBert & 0.811 & 0.845 & 0.828 \\
        BlueBert & 0.807 & 0.860 & 0.833 \\
        PubMedBert & 0.805 & 0.842 & 0.823 \\
        \bottomrule
    \end{tabularx}
\end{table}

In this section, we present the evaluation results of our proposed approach and analyze the performance of various model configurations.

\subsection{Evaluation on Development Dataset}

\subsubsection{Overall Result} Table~\ref{tab:main_results} summarizes the performance of different methods on the development dataset. In this experiment, we utilized GPT-5-mini as the backbone model. As shown in the table, incorporating schemas alongside few-shot examples yields superior performance in extracting automatic medical observations from nurse dictations, compared to providing few-shot examples alone.

\subsubsection{Ablation on Retrieval Model} We conducted an ablation study on the retrieval models used to fetch schemas and few-shot examples. Table~\ref{tab:abl} summarizes the performance results. As shown in the table, BlueBERT achieves the best performance in terms of F1-score.

\subsection{Evaluation on Test Dataset}

In this section, we present the evaluation results on the MEDIQA-SYNUR test dataset, summarized in Table~\ref{tab:test_ex}. Regarding model configuration, the combination of GPT-5.1 as the observation generator and GPT-5-mini as the segmentor yielded the best performance. We also observed a positive correlation between the number of few-shot examples and overall performance. Interestingly, while larger generator models generally improved performance (e.g., GPT-5.1 outperformed GPT-5-mini), this trend was not strictly monotonic. Specifically, GPT-5.2 exhibited performance degradation compared to GPT-5.1. Furthermore, the superior performance of the GPT-5-mini segmentor can be attributed to model consistency. Since the memory bank of segments was constructed using GPT-5-mini, employing the identical model for segmentation during inference ensures better alignment with the retrieved examples, leading to optimal results.
\section{Discussion}

\paragraph{Efficiency in Observation Extraction Pipeline} In this study, we utilized in-context learning by retrieving medical terms from few-shot examples and an ontology. However, during the inference phase, when dictation is divided into segments, not all segments necessarily contain meaningful medical observations. Although this aspect was outside the scope of our current work, we propose that the pipeline's efficiency could be significantly improved by incorporating a preliminary classifier. This classifier would determine the presence of medical observations within a segment before prompting the LLM, thereby reducing unnecessary computational costs for irrelevant segments.

\paragraph{Dataset Analysis} Our analysis indicates opportunities to improve data consistency. For example, values such as 37.5$^{\circ}\mathrm{C}$ sometimes appear as ``375'', and unit annotations (\textit{e.g., }Temperature Unit) are not always consistent. Standardizing value formats and units would likely improve training efficiency and retrieval accuracy.

\section{Conclusion}

In this paper, we proposed a Retrieval-Augmented Generation (RAG)-based pipeline designed to automatically extract medical observations from nurse dictations. To achieve this, we constructed a memory bank leveraging both a medical observation ontology and previously annotated observation tags from existing dictations. By integrating these dataset into the LLM generation process, our approach guides the model to produce more accurate outputs. Consequently, our method demonstrated its effectiveness by achieving an F1-score of 0.796 on the MEDIQA-SYNUR test dataset.

\section{Acknowledgements}

This work was supported by the Korean Government through the grants from IITP (RS-2021-II211343, RS-2022-II220320, RS-2025-25442338).



\bibliographystyle{lrec2026-natbib}
\bibliography{lrec2026-example}

\label{lr:ref}
\bibliographystylelanguageresource{lrec2026-natbib}
\bibliographylanguageresource{languageresource}





\end{document}